\documentclass[conference]{IEEEtran}
\usepackage{graphicx} 
\usepackage{soul}
\usepackage{xcolor}
\usepackage{amsmath}
\usepackage{xspace}
\usepackage{hyperref}
\usepackage{amsthm}

\usepackage{amsfonts}
\usepackage{tabularx}
\usepackage{vcell}
\usepackage{array}
\usepackage{gensymb}
\usepackage{textcomp}
\usepackage{makecell}

\title{Assuring the Safety of Reinforcement Learning Components: AMLAS-RL}

\author{\IEEEauthorblockN{Calum Corrie Imrie\IEEEauthorrefmark{1},
Ioannis Stefanakos\IEEEauthorrefmark{1},
Sepeedeh Shahbeigi\IEEEauthorrefmark{1},
Richard Hawkins\IEEEauthorrefmark{1}, and
Simon Burton\IEEEauthorrefmark{1}}
\IEEEauthorblockA{\IEEEauthorrefmark{1}\textit{Department of Computer Science,
University of York, York, U.K.}\\
Email: \{firstname.lastname\}@york.ac.uk}}

\newcommand{\AMLAS}{\emph{AMLAS}\xspace}

\newcommand{\AMLASRL}{\emph{AMLAS-RL}\xspace}

\newcommand{\artefact}[1]{\textbf{\textsf{[#1]}}}

\begin{document}

\maketitle

\begin{abstract}
    The rapid advancement of machine learning (ML) has led to its increasing integration into cyber-physical systems (CPS) across diverse domains. While CPS offer powerful capabilities, incorporating ML components introduces significant safety and assurance challenges. Among ML techniques, reinforcement learning (RL) is particularly suited for CPS due to its capacity to handle complex, dynamic environments where explicit models of interaction between system and environment are unavailable or difficult to construct. However, in safety-critical applications, this learning process must not only be effective but demonstrably safe. Safe-RL methods aim to address this by incorporating safety constraints during learning, yet they fall short in providing systematic assurance across the RL lifecycle. The \AMLAS methodology offers structured guidance for assuring the safety of supervised learning components, but it does not directly apply to the unique challenges posed by RL. In this paper, we adapt \AMLAS to provide a framework for generating assurance arguments for an RL-enabled system through an iterative process; \AMLASRL. 
    We demonstrate \AMLASRL using a running example of a wheeled vehicle tasked with reaching a target goal without collision.
\end{abstract}


\section{Introduction \label{sec:introduction}}
Reinforcement learning (RL) is a machine learning (ML) technique focused on training \textit{agents}
to generate optimal policies for interacting with their operating environment~\cite{DBLP:books/lib/SuttonB98}.
Inspired by behavioural psychology, RL mimics the human trial-and-error process. An agent takes actions within an environment, observes the resulting outcomes, and learns from feedback in the form of a reward signal. Over time, the agent discovers optimal strategies to achieve long-term objectives. 

RL excels in complex, uncertain, or changing environments, often where human decision-making becomes infeasible due to high dimensionality or delayed consequences. This makes it especially valuable in cyber-physical systems (CPS), where intelligent decision-making must be integrated with real-time control of physical processes. 
Applications in domains such as smart energy systems~\cite{XIONG20243501} and autonomous driving~\cite{kiran2021} can benefit from RL's capacity for self-directed learning, adaptability, and long-term optimisation.  
While traditional ML models often require extensive annotation, RL agents learn by interacting with simulated or real environments, reducing manual effort~\cite{li2018}. Furthermore, RL’s ability to integrate human feedback, when available, makes it flexible for incorporating domain expertise. 

Despite its strengths, RL also presents challenges that limit its widespread deployment in real-world CPS. A major limitation is the high cost or risk associated with real-world experimentation, including domains such as robotics, healthcare, or autonomous driving, trial-and-error learning can result in unsafe behaviours.
This makes simulation-based training a critical component in RL research, allowing agents to learn in virtual environments before being deployed in the real-world~\cite{10.1007/s10994-021-05961-4}. Another significant challenge lies in the interpretability and transparency of RL agents. 
Many RL algorithms, particularly deep-RL, learn complex, non-linear policies encoded in neural networks, making it difficult to understand or justify specific decisions. This ``black-box'' nature raises concerns in safety-critical applications and can undermine trust in deployed systems~\cite{Puiutta2020ExplainableRL}. As a result, developing explainable RL remains an active research area focused on producing models whose decision-making processes are more transparent and understandable to humans.


To address the assurance challenges unique to RL, we adapt the Assurance of Machine Learning in Autonomous Systems (\AMLAS) framework~\cite{hawkins2021guidance}
for RL components. This adaptation, referred to as \AMLASRL, introduces a structured, iterative methodology for generating assurance arguments tailored specifically to RL-enabled components. \AMLASRL accounts for distinctive RL characteristics, such as policy optimisation, state-action feedback loops, and reward-driven behaviour, across six defined stages spanning from safety scoping to model deployment.

This paper presents three key contributions. First, we extend the \AMLAS methodology to support RL applications by defining \AMLASRL as an assurance framework that integrates seamlessly into RL development lifecycles. Second, we detail each stage of \AMLASRL, showing how core assurance activities, 
are adapted to reflect RL’s interactive and dynamic nature. Finally, we demonstrate the practical applicability of \AMLASRL through a case study involving a simulated autonomous vehicle tasked with reaching a goal while satisfying key safety requirements: avoiding energy depletion, limiting time in unsafe zones, and preventing collisions with obstacles.

\section{Background \label{sec:background}}

\subsection{Reinforcement Learning}

RL problems are commonly modelled using Markov Decision Processes (MDPs)~\cite{Uther2010}, which provide a mathematical framework for sequential decision-making under uncertainty. This section formalises the MDP model and introduces key theoretical constructs that underpin the learning and optimisation processes in RL.


The behaviour of an MDP is governed by a \textit{policy}, which resolves nondeterminism by selecting an action to execute in each state. Policies can be classified according to the information they use: a \textit{memoryless} (or \textit{Markov}) policy bases its decision solely on the current state, a \textit{finite-memory} policy uses a bounded portion of past observations, and an \textit{infinite-memory} policy may consider the entire history of interactions.

A policy $\pi$ defines the agent's behaviour as a mapping from states to probability distributions over actions. In the memoryless case, this is represented as $\pi: S \rightarrow \mathsf{Dist}(A)$, where $\mathsf{Dist}(A)$ denotes the set of probability distributions over the action space $A$. The performance of a policy is typically evaluated using \textit{value functions}. The state-value function $V^\pi(s)$ represents the expected return when starting from state $s$ and following policy $\pi$, while the action-value function $Q^\pi(s,a)$ gives the expected return after taking action $a$ in state $s$ and subsequently following $\pi$. The goal in RL is often to compute an \textit{optimal policy} $\pi^*$ that maximises the expected cumulative discounted reward over time.

While standard RL aims to maximise expected return, it does not inherently account for the safety of intermediate decisions or the potential risks associated with exploration. Safe-RL~\cite{10675394} addresses this limitation by integrating safety constraints into the learning process. These constraints are often formalised using cost functions, risk metrics, or probabilistic guarantees, and the objective becomes to learn a policy that not only performs well but also satisfies specified safety criteria.

A common approach is to augment the MDP framework into a constrained Markov Decision process (CMDP), where the agent must optimise its return while ensuring that expected cumulative costs remain below a predefined threshold. Alternative formulations involve risk-sensitive objectives, such as minimising the variance of returns or bounding the probability of catastrophic failures. Safe-RL is particularly critical in domains where unsafe actions can have serious consequences, such as autonomous driving~\cite{9294262}, healthcare~\cite{yu2021reinforcement}, and robotics~\cite{brunke2022safe}. It often relies on techniques like reward shaping, constrained optimisation, shielding, or simulation-based pretraining to balance exploration with risk mitigation.

\subsection{Safety Assurance of Machine Learning}
\begin{figure}
    \centering
    \includegraphics[width=1.0\linewidth]{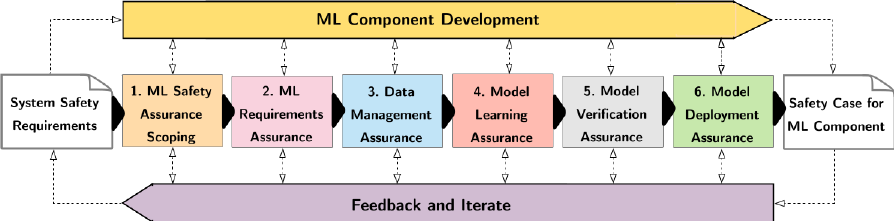}
    \caption{Overview of \AMLAS process.}
    \label{fig:amlas}
    \vspace{-1em}
\end{figure}

Conventional functional safety standards have traditionally failed to address the unique characteristics of ML-based software. Salay et al. \cite{salay2017analysis} analysed ISO 26262—the functional safety standard for road vehicles—and proposed adaptations to better support the integration of ML technologies. Burton et al. \cite{burton2017making} investigated the challenges in assuring the safety of ML-based highly automated driving systems, proposing a contract-based framework to demonstrate the fulfilment of safety-related guarantees under defined assumptions. There are even approaches specifically for RL on how to adopt a safety-first approach, that guides the training of the agent to produce trustworthy systems~\cite{haider2024can, jia2021safety}. 
More recently, the \AMLAS framework provides structured guidance across the ML lifecycle and supports the development of assurance cases for ML components (see Fig.~\ref{fig:amlas} for an overview of the stages). \AMLAS emphasises the importance of an iterative process involving diverse stakeholders and stresses that safety arguments are only meaningful when grounded in the broader system and its operational context.

\section{Motivating example \label{sec:motivating-example}}
\begin{figure}
    \centering
    \includegraphics[width=0.8\linewidth]{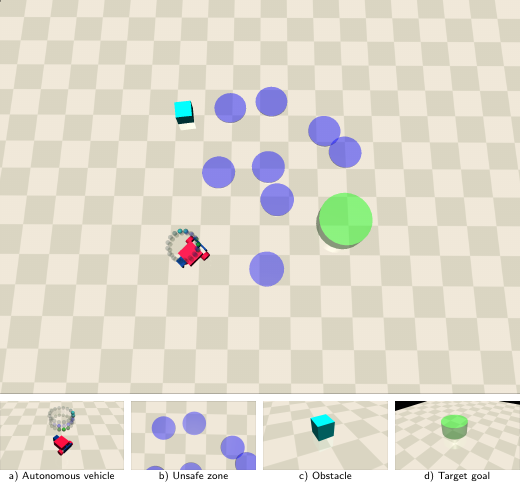}
    \caption{The autonomous vehicle scenario, where the vehicle must reach a goal, minimise time spent in unsafe zones, and avoid collisions with the obstacle.}
    \label{fig:scenario}
    \vspace{-1em}
\end{figure}

An autonomous differential wheeled vehicle is tasked with delivering goods to a goal destination within the environment, see Fig.~\ref{fig:scenario}. Once arrived, there might be additional tasks for the vehicle or the larger system, however, the component to be designed is only concerned with navigating to the goal destination. The vehicle's and goal's position will be denoted as $x_{t}=(x_{t,0}, x_{t,1})$ and $g=(g_{t,0}, g_{t,1})$ respectively. The vehicle can set a  desired velocity at each timestep $t$, $v_{t}=(v_{t,l}, (v_{t,r})$ where $l$ and $r$ denotes the individual velocity commands for the left and right wheels respectively. The wheels have minimum and maximum velocities; $v_{t,i} \in (-1, 1)$.

Within the environment there are eight unsafe zones which will induce risk to the vehicle, denoted as $u=(u_{0},...,u_{8})$, where $u_{i}$ is the coordinates of a specific unsafe zone. These unsafe zones will be considered as a general concern for our purposes rather than anything specific, such as radiation and/or magnetism presence which could damage the carried goods. While being inside an unsafe zone might not cause immediate issues, the vehicle should minimise time being in an unsafe zone. An obstacle is also present, $o=(o_{0}, o_{1})$, and if the vehicle collides with the obstacle it will cause substantial damage causing the mission to end and fail. The initial positions of the vehicle, target goal, unsafe zones, and obstacle are random. 

The vehicle is equipped with three sets of range sensors: one for the target goal ($S_{G,t}$), one for unsafe zones ($S_{U,t}$), and one for obstacles ($S_{O,t}$). Each set consists of 16 range sensors uniformly distributed to provide 
$360\degree$ coverage, i.e., at 
$22.5\degree$ increments. The vehicle has a finite energy supply, $E$, which depletes at a constant rate of $\frac{dE}{dt} = -1$, regardless of the action taken. The vehicle must reach the goal before the energy is fully depleted.

\section{AMLAS-RL\label{sec:amlas-rl}}
Each stage of \AMLASRL is described in this section, along with the running example demonstrating how each stage is executed. While each stage will be presented, this paper will not delve into all the details of \AMLASRL, and instead focus on aspects where \AMLAS has been adapted for RL components. Table~\ref{tab:amlas_rl} lists the inputs and outputs for each stage of \AMLASRL, and which parts are new adaptations from \AMLAS (highlighted in blue). It should be noted that not all inputs and outputs have been included in Table~\ref{tab:amlas_rl} for brevity purposes. 
These include the final output from each stage which is an argument produced from instantiating the stage's corresponding argument pattern with the evidence collected at that stage. The collection of evidence is used to support the instantiated arguments, and both are then key aspects for developing a safety case. As shown in Fig.~\ref{fig:amlas}, the \AMLAS process may necessitate returning to a previous stage, for example a requirement could have been found unsatisfied during stage 5 Model Verification, requiring that previous stages such as data management must be revisited to ensure the requirements can be met. We refer the reader to the \AMLAS document~\cite{hawkins2021guidance} for full details.

\begin{table*}[ht]
    \centering
    \caption{\AMLASRL stages, where blue highlights the required changes and/or considerations when adapting from \AMLAS.}
    \newcolumntype{F}{>{\centering}X}
\newcolumntype{L}{>{\hsize=1.25\hsize}X}
\newcolumntype{S}{>{\hsize=0.5\hsize\centering}X}
\begin{tabularx}{0.85\textwidth}{S|L|L}
\textbf{Stage} & \centering \textbf{Input} & \centering \textbf{Output} \tabularnewline
\hline
\hline
RL Safety Assurance Scoping &
\makecell[l]{
    \artefact{A} System safety requirements\\
    \artefact{B} Description of operating environment of system\\
    \artefact{C} System description\\
    \artefact{D} RL component description
} &
\makecell[l]{
    \artefact{E} Safety requirements allocated to RL component
} \tabularnewline
\hline

RL Safety Requirements Assurance &
\makecell[l]{
    \artefact{E} Safety requirements allocated to RL component
} &
\makecell[l]{
    \textcolor{blue}{\artefact{H} RL safety requirements}
} \tabularnewline
\hline

Data Management & 
\makecell[l]{
    \textcolor{blue}{\artefact{H} RL safety requirements}
} &
\makecell[l]{
    \artefact{L} Data requirements \\ 
    \artefact{M} Data requirements justification report \\
    \textcolor{blue}{\artefact{N} Training plan} \\
    \textcolor{blue}{\artefact{O} Internal test plan} \\
    \textcolor{blue}{\artefact{P} Verification plan}
}
\tabularnewline
\hline

Model Learning &
\makecell[l]{
    \textcolor{blue}{\artefact{H} RL safety requirements} \\
    \textcolor{blue}{\artefact{N} Training plan} \\
    \textcolor{blue}{\artefact{O} Internal test plan} 
} &
\makecell[l]{
    \textcolor{blue}{\artefact{U} RL model development log} \\
    \artefact{V} RL model \\
    \artefact{X} Internal test results
} \tabularnewline
\hline

Model Verification &
\makecell[l]{
    \textcolor{blue}{\artefact{H} RL safety requirements} \\
    \textcolor{blue}{\artefact{P} Verification plan} \\
    \artefact{V} RL model
} &
\makecell[l]{
    \artefact{Z} RL verification results \\
    \artefact{AA} Verification log 
} \tabularnewline
\hline

Model deployment &
\makecell[l]{
    \artefact{A} System safety requirements \\
    \artefact{B} Description of operating environment of system \\
    \artefact{C} System description \\
    \artefact{V} RL model \\
    \textcolor{blue}{\artefact{EE} Operational scenarios}
} &
\makecell[l]{
    \textcolor{blue}{\artefact{DD} Erroneous behaviour log} \\
    \artefact{FF} Integration testing results 
}
\end{tabularx}
    \label{tab:amlas_rl}
    \vspace{-1em}
\end{table*}

\subsection{Stage 1 RL Assurance Safety Scoping}
The scoping stage in AMLAS sets the boundaries for safety assurance of the RL component and defines the top-level safety claim, supported by key contextual inputs: description of the operating domain~\artefact{B}, system description~\artefact{C}, and ML component description~\artefact{D}. These inputs guide the allocation of safety requirements to the RL component~\artefact{E} which must be technology-agnostic and ensure safe integration within the system. While defining these requirements is part of the broader system safety process, \AMLAS assumes their correctness and requires evidence in the overall safety case. This stage of \AMLASRL is the same as defined by the original \AMLAS framework.

\medskip
\noindent\textbf{Safety scoping for vehicle example.}
The development cycle begins with understanding the purpose of the RL component, along with the operating domain of the vehicle in the environment. Using domain knowledge, it is determined that the vehicle will need to complete navigational mission with a $60\%$ minimum success rate, and will have an initial energy value of 250. As part of this stage, HAZOP was undertaken resulting in identification of being inside an unsafe zone, cumulatively, for around 20 timesteps or more can lead to substantial risk to the carried good. Lastly, while vehicles can potentially break down, they can be recovered, but recovering them if the vehicle is damaged creates risk to the operators tasked with retrieval. All of these considerations lead to the following safety requirements. 
\begin{itemize}
    \item \textbf{SR1}: The vehicle will reach the destination area with $E>0$ with a probability of at least 0.6
    \item \textbf{SR2}: The vehicle will not be in an unsafe zone for longer than 20 timesteps
    \item \textbf{SR3}: The vehicle will not collide with the obstacle on any mission with probability of 0.1
\end{itemize}

\subsection{Stage 2 RL Safety Requirements Assurance}

This stage of \AMLAS methodology defines a set of ML safety requirements~\artefact{H} for the ML component. These requirements are translations from the safety requirements~\artefact{E}, identified at Stage 1, to a form that will enable engineers to implement the RL component correctly such that the system level requirements can be satisfied. An example would be understanding of the consequences of false positives and false negatives for an object detection unit, and what are the acceptable levels of inaccuracy to meet the safety requirements. 

RL perceives a state to then generate and, subsequently, enact an action. Consequently, RL is further entwined intrinsically with the system's overall behaviour compared with supervised learning. The relationship in the state-action space is also usually unknown to the stakeholders, which is what the RL agent has been tasked with understanding and exploiting. This makes it harder to define ``safe" behaviour to compare directly with individual state-action pairs, and instead should be analysing the sequence of steps to understand the overall behaviour. Therefore, the RL safety requirements will overlap with the system level requirements. However, this overlap will be dependent on the level of control the RL component exerts onto the system, for example the differences between a car fully controlled and partially controlled, such as steering, by an RL controller.

The RL's learning will be guided by the reward function, and would be constructed to incorporate all aspects of the desired behaviour to achieve the intended goal. This is a useful start point for identifying if the engineers would have difficulty with any of the safety requirements when attempting to encode them into the reward function. In Safe-RL, for example, special consideration is given to violations, with multiple approaches on how to inform the RL agent of these violations. It can be something as simple as including violations as a penalising cost signal into the reward function, or designing the architecture as a CMDP. The latter utilises an additional reward function only for the costs of violations with dynamic weighting between the different reward functions. While the model is implemented at a later stage in \AMLASRL, the identification of these violations are useful for the RL engineering team. Primarily because it improves clarity on what tradeoffs will exist, as the costs may oppose behaviours which optimises achieving the desired goal.


\medskip
\noindent\textbf{Safety requirements for vehicle example}.
The RL controller has complete control of the autonomous vehicle, therefore the overlap from the previous stage will be absolute. The requirements from Stage 1 are all behavioural concerned, and the SRs are sufficiently descriptive for the RL engineer, and therefore requiring no additional translation in this instance. Additionally, SR2 and SR3 are direct safety violations that contradict the overall aim of reaching the goal. These are highlighted as such and passed onto the RL engineering team. While SR1 is a safety violation it does not oppose the main objective of reaching the target goal. 

\subsection{Stage 3 Data Management Assurance}
An ML model's performance will be heavily reliant on the data that is used for training. This stage of \AMLAS is to develop data requirements that are in accordance with the safety requirements derived \artefact{H}, along with a report to justify the sufficiency of those data requirements~\artefact{M}.  For \AMLAS, the data requirements considers the following properties~\cite{ashmore2021assuring}:
\begin{itemize}
    \item \underline{Relevance}: how does the data match the operating domain;
    \\\textit{``the same camera model should be used for collecting data samples as the one that is deployed on the system''}
    \item \underline{Accuracy}: is the data correct with respect to ground truth;
    \\\textit{``images of benign and cancerous skin lesions for the melanoma ML classifier must be checked by a minimum of three experts, and if in doubt should label as cancerous''}
    \item \underline{Balance}: is the distribution of the sampled and/or generated data appropriate;
    \\\textit{``the all-year weather forecast ML predictor must have as close to an equal number of data samples from all 12 months, with the tolerance dependent upon the number of days in each month''}
    \item \underline{Completeness}: what is the data's coverage of the operating domain;
    \\\textit{``the outdoor security camera expected to operate continuously, the dataset must contain both day and night images of the area covered by the camera''}
\end{itemize}

The examples provided with each property are from the context of supervised learning, and the data requirements would be fully explored during this stage for \AMLAS. While the specifics may change of what is necessary to properly understand the data and requirements with respect to RL, the properties are still applicable. For each problem there will be nuances that will influence how the property is used to consider and generate requirements, however the state-action space will be a primary factor for the properties. Therefore how the states are generated and exposed to the RL agent are important for the data requirements, as this will ensure effective exploration of the state-action space.

To highlight how the properties can be applied to RL agents, let us consider the usage of a simulator. Training RL agents usually requires massive state-action observations to learn policies. It is therefore typical practice when building RL agents for physical systems to perform the training phase within a simulator to conserve time and resources. The simulation is providing data to the RL agent to learn a policy to then be deployed in the physical environment. Essentially, the simulator is a tool for generating states and the subsequent states from state-action pairs. If a simulator is used, it must replicate real-world dynamics (accuracy) and adequately cover the operating environment (completeness). Further, how the simulated input data to generate the RL agent's state needs to mirror the expected deployed system (relevance). The training plan (detailed later) will also need to be formulated to encourage exploration and revisit states where appropriate (balance). 

The high level examples can be applied beyond simulators. Additional learning might be performed for physical systems post simulation with the physical system itself, as fine tuning may be required due to a simulation-reality gap. The physical set up for the training is still generating states, and if this to be part of the training then consideration must be taken with respect to the properties. 

In \AMLAS, the data requirements are then used for generating datasets for the subsequent stages of model learning and verification, but this is not directly applicable for RL. This stage will now instead produce a training plan \artefact{N}, an internal test plan \artefact{O}, and a verification plan \artefact{P}. 

The training plan is how the RL agent will learn the desired behaviour, which might include choice of simulator, process of executing and concluding learning episodes, design decisions for implementing state perception and action execution, etc. The training plan should also include insight of how the development team will track and record the learning progress. It would be expected for the reward function to be part of this, though the design of the reward function will occur in the next stage, Model Learning Assurance. Therefore, the reward function should be assumed to provide minimal information of the RL agent's overall behaviour and performance, i.e. whether the agent is succeeding or failing to learn, and if it is or has converged. The internal test plan will detail how the development team will evaluate their learned model, which might include general and/or targetted tests. Lastly, an independent verification team creates the verification plan to evaluate the final RL model. The verification plan uses the data requirements to define expected behaviours and differs from the internal test plan, as it focuses on identifying if and when the RL agent fails during deployment. The development team does not have access to the verification plan while developing the system. The training and internal test plans are used in stage 4, while the verification plan is applied in stage 5.


At this point, it is worth reiterating a key strength of \AMLAS: the generation of assurance arguments at crucial stages of the development cycle. By following \AMLASRL, the process ensures that system and ML safety requirements are established before model construction. These requirements inform the data requirements, which in turn guide model implementation. If a requirement is found to be unsatisfied, \AMLASRL can be repeated, but now with the existing artefacts and insights to re-evaluate earlier stages. This iterative approach helps avoid wasted effort, such as implementing and training an RL model only to later discover that the requirements are inadequate, the agent's capabilities are insufficient, or the system's operating domain is too broad to ensure safety.

\medskip
\noindent \textbf{Constructing training, testing, and verifying plans for the vehicle example.}
The vehicle's dynamics and the environment it will be operating in can be perfectly captured using Safety-Gymnasium. It provides the capability of encapsulating the operating domain by simulating the 3 sets of lidar sensors, the vehicle dynamics, and the number and placements of the unsafe zones, obstacle, and target goal. 

The location of each object can be any unoccupied position in the environment. Accordingly, the training plan specifies that object placement for each episode will be uniformly random. The goal of training is to learn a suitable policy for reaching the goal location while avoiding obstacles, as collisions are considered catastrophic. Each episode will terminate when the vehicle either reaches the goal or collides with an obstacle. Additionally, to ensure sufficient exploration of the state-action space, the episode will also terminate after 500 timesteps. To support exploration, this value is allowed to exceed the energy limit specified in \textbf{SR1}.

For each episode, the development team will record the following: total episode reward, whether the goal was reached and the time taken (if successful), any obstacle collisions, and time spent in unsafe zones. Although the reward function will influence the overall training time, the team, based on prior experience, will initially train for 500 episodes. Monitoring the timecourse of these tracked metrics will provide insight into learning convergence and behavioural improvement. This, in turn, will determine whether the RL model is ready for testing, requires further training, or needs to be reimplemented.

The internal test plan will consist of 1000 trials, each with randomly placed objects. Trials will terminate when the vehicle reaches the goal, collides with an obstacle, or depletes its energy. Metrics will be recorded across trials, and average outcomes will be compared against defined requirements.

The verification team are aware of the safety and data requirements. The verification team will initially test the trained vehicle with 500 random trials to evaluate for the same purposes, but additionally will test for 250 trials where the vehicle begins near an obstacle; between 0.2 and 0.3 metres.
The verification team will also create a discrete time Markov model (DTMC) for model checking, via abstraction of the complete state space. Specifically, let a distance function for the vehicle at timestep $t$ be defined as $d(y)=\sqrt{(x-y)^{T}(x-y)}$ where $y$ is the positions of an object in the environment. Let a state at timestep $t$ be

{\small
\begin{equation}
    s_{t}=(m_{t},e_{t}, d_{o,t}, d_{u,t})
\end{equation}}

\noindent where $m$ is the current representation of the vehicle's journey; travelling inside/outwith of an unsafe zone, collided with the obstacle, or reached the goal.

{\small
\begin{equation}
    m=
    \begin{cases}
        0 & \text{if } d(g_t)>0 \\
        1 & \text{if } d(g_t)>0 \land \exists \;u_{t} \in U_{t}.  d(u_{t})<\hat{u} \\
        2 & \text{if } d(o_t)<\hat{o} \\    
        3 & \text{if } d(g_t)=0 \\
    \end{cases}
\end{equation}}

\noindent The variable $e_{t} \in [0, E]$ is the energy remaining, and the values $\hat{u}$ and $\hat{o}$ are thresholds representing the size of the unsafe zones and obstacle; these are 0.2 and 0.1 respectively.


\noindent The transition probabilities will be extracted from running 5000 trials. State $s_t$ is considered terminal if $m_{t}\in\{2,3\} \lor e_{t}=0$. The verification team will use probabilistic computational tree logic (PCTL) to encode the safety requirements as properties for model checking

{\small
\begin{equation}
    \begin{split}
        C1 &: \mathcal{P}[m=2]\ge 0.6 \\
        C2 &: \mathcal{P}[m=3]\le 0.1 \\
        R0 &: \mathcal{R}^{\textsf{unsafe}}[\text{ }\mathrm{U}\text{ } m=2]    
    \end{split}
\end{equation}}

\noindent $C1$ evaluates \textbf{SR1} (as $m=2$ can only occur while $e>0$), and $C2$ evaluates \textbf{SR3}. $R0$ is a reward based property that tracks how often the vehicle is within an unsafe zone. These will be performed with the PRISM model checking software. 

\subsection{Stage 4 Model Learning Assurance}
Here the RL model is constructed~\artefact{V}, and is trained as guided by the training plan~\artefact{N}. This includes formulating the reward function. The RL model produced is utilised in a similar manner to other ML models, hence why it is not coloured blue in Table~\ref{tab:amlas_rl}. 

The design choices are recorded as part of a RL model development log~\artefact{U}, which would include the model's architecture, libraries used, etc. Given the importance of the reward function, it is paramount that this is included in the log as well, along with a cost function if employed. The problem comprehension would improve, and thus impart insight if the development cycle requires returning to an earlier stage. For example, RL agents can be subjected to reward hacking~\cite{skalse2022defining}, which is when the policy that optimises the reward function produces undesirable behaviours. In this instance, it might be discovered that the RL safety requirements are lacking sufficient details or coverage to properly inform the development team how to create a suitable reward function. 

Multiple candidate RL models can be constructed, and they do not need to use the same reward function. The development team then evaluates the model(s) using the internal test plan~\artefact{O} to check how well the RL safety requirements~\artefact{H} are satisfied. From the internal test results~\artefact{X} the best candidate RL model will be identified and outputted. If, though, no RL model satisfies the requirements then process will not be able to continue, and the development may need to return to an earlier stage of \AMLASRL.  

\medskip
\noindent \textbf{Implementing RL models for vehicle example.}
The RL model will use the range sensor data as the input state, size 48, and output velocity control to the wheels as the action, size 2. The development team has decided to use a Deep Deterministic Policy Gradient (DDPG) architecture, with the justification being that DDPG are an appropriate option for continuous state and action spaces. DDPG consists of two networks; a critic network which learns the value function from the rewards observed from experienced state-action pairs, and an actor network which learns the best policy using the value returned from the critic network. Both networks are feed forward, with the critic being a multi-input network, an input vector each for the state and action, and an input vector for the actor containing the state. Full details of the models can be found in the \href{https://github.com/ccimrie/SEAA\_AMLAS-RL}{github repository}.

The reward function is designed to reward the agent if it reduces its distance to the goal. To deter the vehicle from colliding with the obstacle and being in unsafe zones, a penalising cost value is applied 

\begin{figure}
    \centering
    \includegraphics[width=0.9\linewidth]{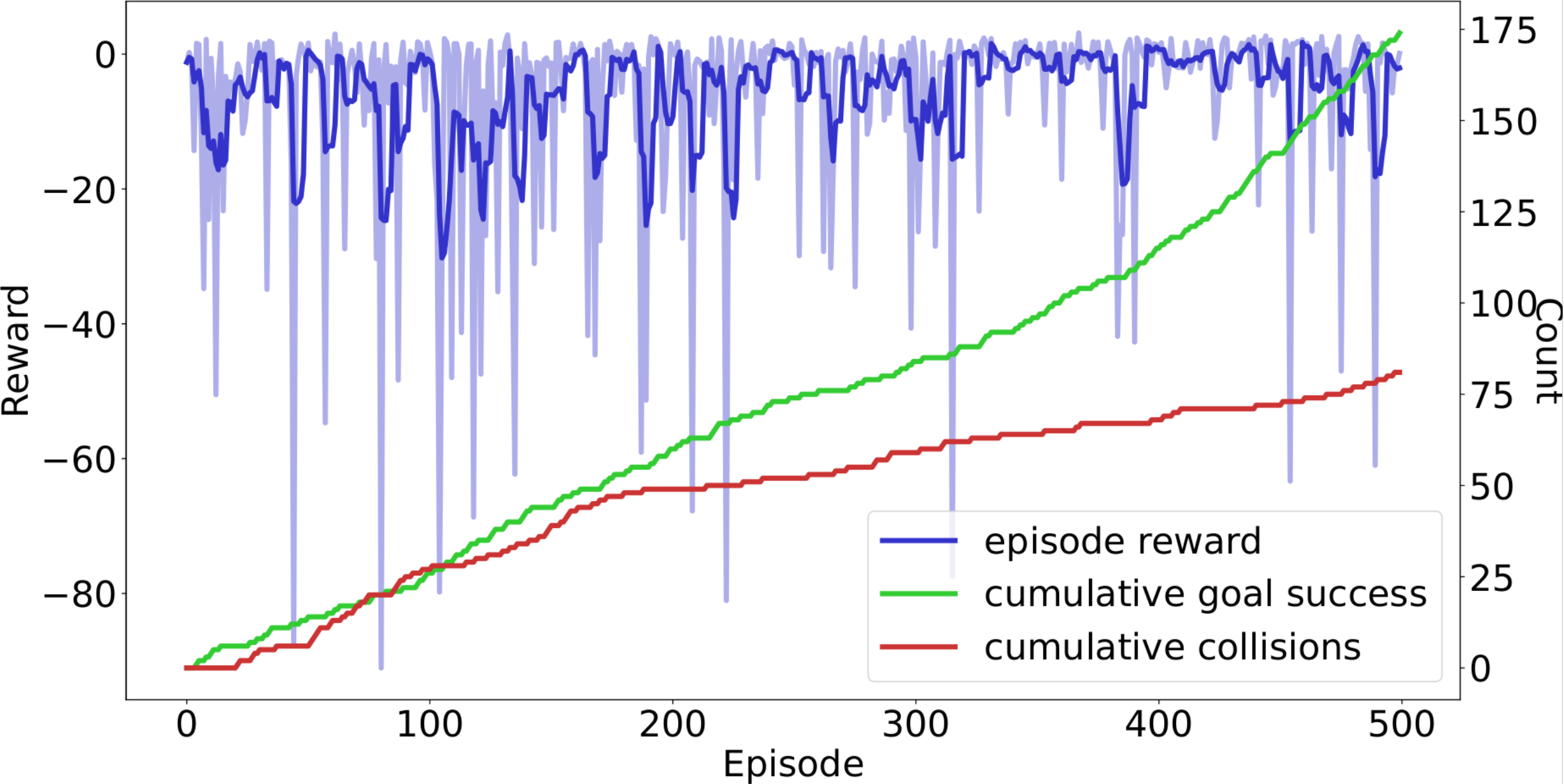}
    \includegraphics[width=0.9\linewidth]{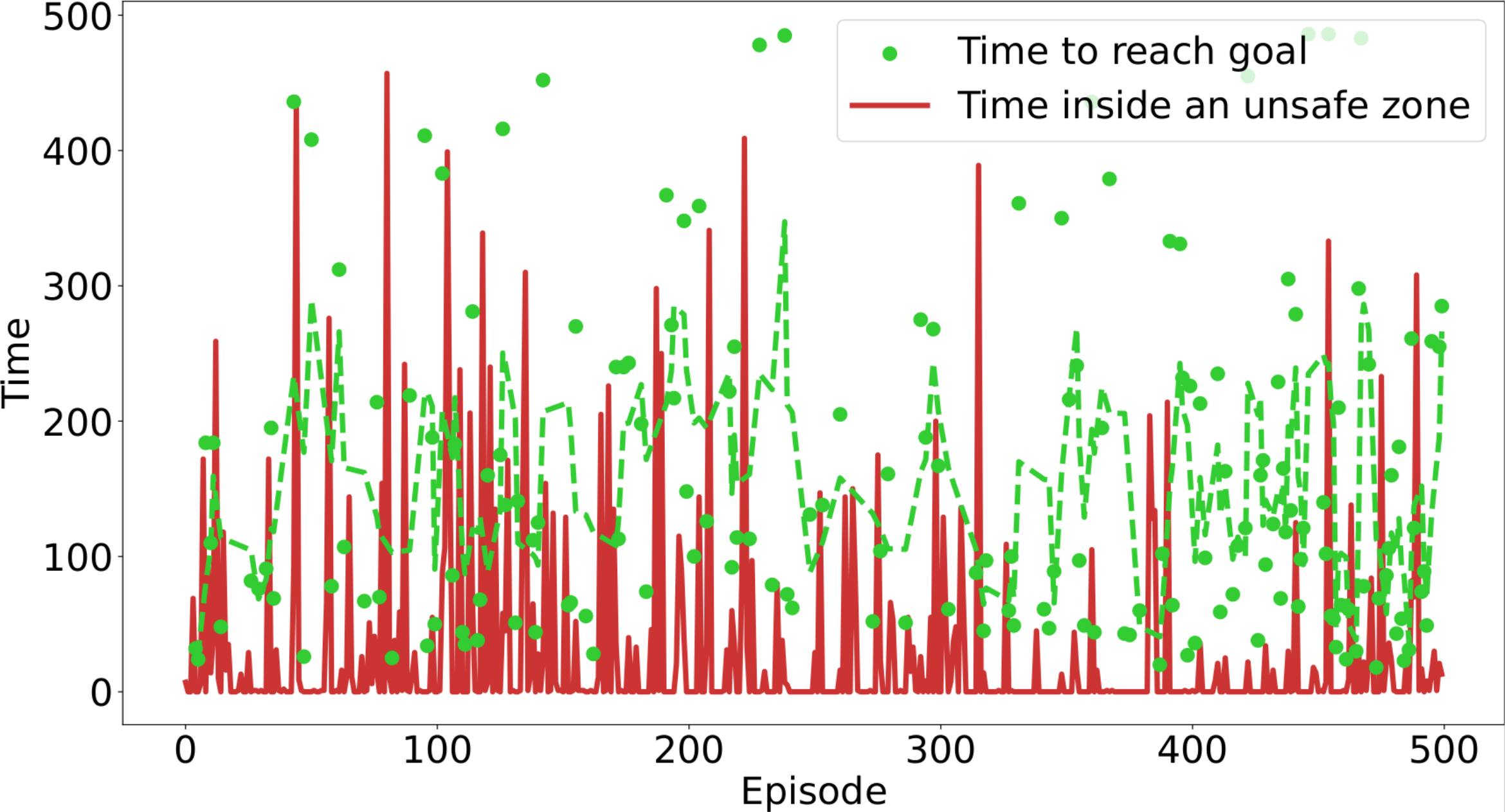}
    \caption{Plots of the RL's training. Top plot displays the total reward, cumulative goal-reaching successes, and collisions. Bottom plot shows the time taken to reach the goal (when successful) and time inside unsafe zones.}
    \label{fig:rl_training}
    \vspace{-1em}
\end{figure}

{\small
\begin{equation}
    \begin{split}
        R_{t}&=r_{t}-c_{t} \\
        r_{t}&=(D_{t-1}-D_{t})\beta \\
        c_{t}&= 
        \begin{cases}
            -0.1 & \text{if $\exists u_{t} \in U_{t}.d(u_{t})<\hat{u}$} \\
            -10 & \text{if $d(o_{t})<\hat{o}$} \\
        \end{cases}
    \end{split}
\end{equation}}

\noindent where $\beta$ is a discount factor. Fig.~\ref{fig:rl_training} contains the timecourses for the various values to be tracked as described in the training plan. The team observes the agent has learned a policy, and the number of successful missions continue to increase and collisions significantly less so. 
The development team then tests the trained model with 1000 trials as described in the internal test plan, and recording the results, see Table~\ref{tab:learning_results}. The results indicate satisfaction of the RL safety requirements, and the RL model is passed to the verification team. 

\begin{table}[!t]
    \centering
    \caption{Results on trained agents from tests conducted in Stage 4.}
    \renewcommand{\arraystretch}{1.3}
    \begin{tabular}{|c|c|}
    \hline
    \textbf{Description} & \textbf{Test results}\\
    \hline
    Percentage reaching goal with $e>0$  & $60.2\%$ \\
    \hline
    Average energy remaining during a successful mission & $161$ \\
    \hline
    Average time per mission in an unsafe zone & $15.94$ units \\
    \hline     
    Percentage of collision with the obstacle  & $7.56\%$ \\
    \hline     
    \end{tabular}
    \vspace{-1em}
    \label{tab:learning_results}
\end{table}

\subsection{Stage 5 Model Verification Assurance}
With the internal testing indicating successful learning, the RL model~\artefact{V} is analysed by the verification team with the verification plan~\artefact{P}. These are to ensure that the model complies with the RL safety requirements~\artefact{H}. This stage can utilise appropriate tests and formal methods techniques to evaluate the model, as guided by the verification plan. During this stage, \AMLASRL does not differ much from \AMLAS, with the most notable exception being the verification plan (detailed in stage 3). The core aim is maintained between \AMLAS and \AMLASRL, and this stage highlights the need to extend beyond the reward function when evaluating RL agents. It is not to say the reward function does not have its benefit for understanding if the RL agent is succeeding, as described in Stages 3 and 4. Rather, the verification performed should be targetted at evaluating the RL model in potentially unsafe scenarios which therefore requires metrics that are problem context specific. In particular, scenarios that can identify possible reward hacking issues would be of substantial benefit for informing the development cycle.

While formal methods can be employed, a level of abstraction will mostly likely be required. Once a model has been created, it should be properly assessed as to whether the abstraction process results in a model which is still capable of capturing the important elements of the system. Effort should be made to explore the utilisation of formal methods, as even if a formal model cannot be produced for this Stage to directly assess the RL model's compliance with the RL safety requirements, the very act of creating the formal model encourages a proactive exploration of the problem. Additionally, the formal model could provide insight into the behaviour of the RL model to understand the verification results~\artefact{Z}. For example, a model checking process could reveal that the RL agent favours being close to objects when it traverses a particular trace.  

\noindent \textbf{Verification of RL-controlled vehicle example.}
The verification team follows the verification plan. First, 500 trials are performed with randomised locations, with 250 trials when the agent is initialised near the obstacle, between 0.2 and 0.3 metres. The verification team also conducts 5000 trials to gather data for calculating transition probabilities for the DTMC detailed in stage 3. Table~\ref{tab:verification_results} presents the test and model checking results, which are in agreement of the vehicle's compliance with the safety requirements. The targetted testing, though, demonstrates the vehicle struggling, and when in operating in this situation the RL vehicle is unable to satisfy SR1 and SR2. It is therefore advisable to revisit Stage 3 to re-evaluate the training, internal test, and verification plans. 

\begin{table}[!t]
    \centering
    \caption{Results from Stage 5 of \AMLASRL.}
    \renewcommand{\arraystretch}{1.3}
    \begin{tabular}{|>{\centering\arraybackslash}p{2.5cm}|>{\centering\arraybackslash}m{1.5cm}|>{\centering\arraybackslash}m{1.5cm}|>{\centering\arraybackslash}m{1.5cm}|}
    \hline
     \textbf{Description} & \textbf{General Verification results} & \textbf{Targeted Verification results} & \textbf{Model checking results}\\
    \hline
     Percentage reaching goal with $e>0$  & $60.8\%$ & $57.6\%$ & $60.4\%$ \\
    \hline     
    Average time per mission in an unsafe zone & $15.51$ units & 23.48 timesteps & 17.52 timesteps \\
    \hline     
    Percentage of collision with vase  & $5.4\%$ & $7.6\%$ & $5.8\%$ \\
    \hline     
    \end{tabular}
    \vspace{-1em}
    \label{tab:verification_results}
\end{table}

\subsection{Stage 6 Model Deployment Assurance}
The final stage is observing the integration of the RL model~\artefact{V} into the deployed system, using the description of the operating environment~\artefact{B} and the system description~\artefact{C}. Operational scenarios~\artefact{EE} are  conceived to evaluate whether the system, now with the RL model integrated, satisfies the system safety requirements~\artefact{A}. The outcome is to produce integration testing results~\artefact{FF}. While performing the integration, the \AMLAS methodology also produces an erroneous behavioural log~\artefact{DD} tasked with recording the model's output when failures have been detected. This is to account for the uncertainty that will be present in many ML models. For \AMLASRL the erroneous behavioural log will need to account for sequence of states, and might therefore require analysis techniques that have a longer time horizon. This is useful for identifying the possible presence of reward hacking as well as the general behaviour of the system. 

\medskip
\noindent\textbf{Deploying the RL vehicle.}
The RL vehicle was unable to satisfy the requirements during the verification at Stage 5. If, though, the development cycle upon the repetition of necessary stages produced a model that succeeded in Stage 5 the RL vehicle would still need to be assured in Stage 6. This is, however, part of a larger mission of delivering goods, and the RL model produced is only for the navigation. As stated in Sec.~\ref{sec:motivating-example}, the unsafe zones might impact the integrity of the carried good. The operational scenarios would therefore observe the mission, including the tasks the system performs before and after the RL vehicle traverses the environment. Here, the scenarios would be checking that the uncertainty within the RL model does not affect the quality of the carried goods. Further, the team decides to record the sequence of behaviours from the vehicle, in particular if the agent crashes or enters an unsafe zone. If this proves to fail at this stage, the development cycle may return to Stage 4 and using the erroneous behavioural log identify the immediate state-action pair that led to one of these faults. This may then prompt the team to implement a post-shielding approach; a technique that would override the action generated by the RL model with a predefined action known to ensure safety to the vehicle and carried good. Further, the team decided to observe if undesirable behaviours had been learned as a result of reward hacking. For example, the vehicle may have optimised the reward function, but have learned the policy to always drive in reverse, which depending on how the good is being carried by the vehicle could lead to dangerous consequences. 
\section{Related Work\label{sec:related-work}}
Safe-RL is an active research area that has advanced both the theoretical foundations of RL and the development of algorithmic techniques~\cite{10675394}. Notably, Safe-RL aims to minimise risky or unsafe behaviour by introducing new algorithms, theoretical guarantees, and domain-specific methods. Our work shares this overarching goal of enabling RL in safety-critical CPS, though \AMLASRL addresses the development life cycle. 
This aids the body of Safe-RL research, as many developed techniques and practises can support the aims of \AMLASRL's stages. Fundamentally, \AMLASRL provides a structured, system-level assurance methodology for guiding the development of assurance arguments throughout the RL lifecycle.

The authors of~\cite{safeRLHabli} explore the safety challenges of deploying RL in safety-critical systems, arguing that traditional assurance methods based on deterministic behaviour (Safety-I) are inadequate for adaptive systems using RL. 
They propose a high-level safety argument using Goal Structuring Notation, highlighting safe reconfiguration, fail-safe transitions, continuous monitoring, and a dynamic assurance approach aligned with Safety-II. In contrast, \AMLASRL offers a practical, stage-by-stage adaptation of the \AMLAS methodology tailored to RL. While the former focuses on high-level theory, \AMLASRL introduces a concrete assurance process, with defined stages, safety artefacts, and validation via a case study.

To address the lack of accessible safety cases for ML-enabled autonomous systems, recent work~\cite{10628705} presents a complete safety case for an RL component deployed on the Quanser QCar platform. Building on the \AMLAS and SMIRK frameworks, the methodology adapts them to an existing system, focusing on hazard analysis, ML safety requirements, and deployment validation. 
In contrast, this work applies only three stages of \AMLAS post hoc to a pre-developed RL component on a physical platform. Our approach is proactive, designed for use throughout the RL development lifecycle. \AMLASRL focuses on how RL-specific challenges such as state-action uncertainty, reward structure interpretation, and simulation-to-reality transfer, can be addressed. 

Another line of work~\cite{Niu31122025} introduces the SARL framework for autonomous vehicles, which combines deep RL with a formal safety layer based on the Responsibility-Sensitive Safety model. It integrates mathematically defined safety constraints into the agent’s decision-making, enabling real-time enforcement of safe behaviours during training and deployment. 
In contrast, \AMLASRL offers a more general, lifecycle-oriented assurance approach and provides stage-by-stage guidance for generating assurance arguments, making it applicable to a broader range of CPS beyond autonomous driving.
\section{Conclusion \label{sec:conclusion}}
RL can learn optimal behaviours in complex settings and is well-suited when the state space model is unknown or hard to specify. This makes it appealing for enhancing CPS capabilities. However, as the technology advances, assuring RL components and maintaining system-level safety is essential.
\AMLAS is a methodology for developing a safety case for ML models by generating arguments and evidence from assurance activities and artefacts across the ML lifecycle.
This paper presented \AMLASRL, an adaptation of \AMLAS for RL components. Its application was demonstrated through a running example of an RL-operated vehicle navigating to a goal as part of a goods delivery system. 

Future work includes exploring the potential overlap between stages 3 and 4 resulting from the use of RL. For instance, the reward function, although constructed in stage 4, influences the agent’s exploration of the state space and thus affects the data-related properties defined in stage 3. To investigate this and further evaluate \AMLASRL, the existing case study will be scaled up into a full demonstrator project. While the task remains the same, the demonstrator will use a physical robot (e.g., the Husky) and address the added challenge of generating assurance evidence when training in simulation (e.g., Gazebo) before deployment in the real world.

\bibliographystyle{ieeetr}
\bibliography{references}

\end{document}